\documentclass[oribibl]{svproc}

\usepackage{amsmath}
\usepackage{amsfonts}
\usepackage{amssymb}
\usepackage{graphicx}
\usepackage{bm}
\usepackage{booktabs}
\usepackage{multirow}
\usepackage{makecell}
\usepackage{hyphenat}
\usepackage{url} 

\newcommand{\tp}{^{\mathsf{T}}}
\newcommand{\point}{~.}
\newcommand{\comma}{~,}

\DeclareMathOperator{\diag}{diag}
\DeclareMathOperator{\var}{var}
\DeclareMathOperator{\vecop}{vec}

\begin{document}

\mainmatter              
\title{PCA-Based Interpretable Knowledge Representation and Analysis of \\ Geometric Design Parameters}
\titlerunning{Knowledge Representation and Analysis of Geometric Design Parameters}  
%
\author{Alexander K\"ohler \and Michael Breu\ss} 
\authorrunning{Alexander K\"ohler and Michael Breu\ss} 
%
\tocauthor{Ivar Ekeland, Roger Temam, Jeffrey Dean, David Grove,
Craig Chambers, Kim B. Bruce, and Elisa Bertino}
%

\institute{Brandenburgische Technische Universität\\ Platz der deutschen Einheit 1, 03046 Cottbus, Germany\\
\email{ \{koehlale, breuss\}@b-tu.de} 
}

\maketitle              

\begin{abstract}
	In many CAD-based applications, complex geometries are defined by a high number of design parameters.
    This leads to high-di\-men\-sion\-al design spaces that are challenging for downstream engineering processes like simulations, optimization, and design exploration tasks. 
    Therefore, dimension reduction methods such as principal component analysis (PCA) are used. 
    The PCA identifies dominant modes of geometric variation and yields a compact representation of the geometry.
    While classical PCA excels in the compact representation part, it does not directly recover underlying design parameters of a generated geometry. 

    In this work, we deal with the problem of estimating  design parameters from PCA-based representations. 
    Analyzing a recent modification of the PCA dedicated to our field of application, we show that the results are actually identical to the standard PCA.
    We investigate limitations of this approach and present reasonable conditions under which accurate, interpretable parameter estimation can be obtained.
    With the help of dedicated experiments, we take a more in-depth look at every stage of the PCA and the possible changes of the geometry during these processes. 
\end{abstract}

\section{Introduction}

In modern CAD-based tasks, geometries are often defined by dozens or hundreds of parameters, resulting in high-dimensional design spaces.
Such high dimensionality leads to several challenges for downstream engineering use,
including high computation time, ill-conditioned optimization problems, and highly correlated variables, 
which complicate, e.g., design exploration and simulation. \cite{Cinquegrana2018, Harries2021, Yasong2018}

To address issues related to high dimensionality, dimension reduction methods are commonly employed. 
One widely used method is the principal component analysis (PCA).
Originally, the PCA was introduced by Karl Pearson \cite{Pearson1901} in 1901, and in 1933, independently of Pearson, Harold Hotelling, developed an analogous method to Pearson's \cite{Hotelling1933}. 
The PCA identifies orthogonal modes of variation in data, and provides a compact data representation. 
The simplicity and robustness of the method made it a standard tool in machine learning tasks \cite{Chen2017, Howley2005}, image processing \cite{Bouwmans2018, Patil2011}, genetics \cite{McVean2009, Reich2008}, and finance \cite{Le2020, Nobre2019}.
The book \cite{Jolliffe2002} provides a decent overview on this topic.

To create a connection between the generating parameters and the variations of a geometry bundled into a geometry class, the PCA method is the tool of choice.
It is expected that eigenvectors yielded from the PCA,
which form a latent space,
and eigenvalues that capture the main modes of variation between the elements of the class,
have a relation to the generating parameters. 
This compact representation of geometry class may enable knowledge representation that can be tested by reconstructing an element of the class.
However, a connection between the standard PCA and geometrically interpretable design parameters needs to be established explicitly.

Several recent works try to overcome limitations of PCA by recovering original design parameters, or other related information from the latent spaces. 
The work of Diez et al. \cite{Diez2015} used an approach based on Kosambi–Karhunen–Loève expansion.
In \cite{Serani2025}, Serani et al. refined their dedicated PCA extension \cite{Serani2023}, to include physical data as well.
Something similar was done in \cite{Phuong2025}, where the authors used PCA and inverse neural networks to predict mechanical properties from the dedicated geometry. 

\textbf{Our contribution} is twofold. 
First, we show that parameter estimation results of the extended PCA model in \cite{Serani2023} are actually equal to the ones retrieved by some aftermath of the standard PCA. 
The second contribution would be that we show the limits of interpretable parameter estimation. 
We explore possible conditions that may lead to good parameter estimation results. 
Along the way, we deepen the understanding of what happens to geometries within the PCA. 

The paper is organized as follows. Section \ref{sec:theoretical_background} provides an overview of the main theoretical aspects. 
The notion of a geometry, the theory of both PCA methods, and how to estimate generating parameters from their results are presented there.
Additionally, the connection between both methods is drawn. 
In Section \ref{sec:Experiments}, all experiments are performed. 
First, we introduce a handful of instructive example geometries. 
Then, we follow these geometries on their way through the PCA and show the changes that occur, and the effect of these changes on the estimation quality.
We finalize the paper  with the conclusion in Section \ref{sec:Conclusion}. 

\section{Theoretical Background\label{sec:theoretical_background}}

This section provides an overview of the main theoretical aspects of this paper. 
Starting with the definition of a geometry, moving to an introduction of the standard PCA, and ending with the PCA variant presented in \cite{Serani2023}. 
The methods to extract design parameters from both PCA results are explained along the way. 

\subsection{Notion for Geometries\label{sec:geometries}}
The core of the experiments are geometries $\mathcal{G}\subset \mathbb{R}^3$, which are realizations of a geometrical object $\mathcal{B}$. 
A geometry $\mathcal{G}$ is realized as a compact 1-D or 2-D Riemannian manifold of a 3-D geometric object $\mathcal{B}$.

Furthermore, all geometries are declared as the bounding surface of the object $\mathcal{G} = \partial \mathcal{B}$.
They are represented as point clouds $G \subset \mathbb{R}^3$ with $n\in \mathbb{N}$ points $(x_i,y_i,z_i) = g_i \in G$ for $i\in\lbrace 1,2, \ldots, n \rbrace$.
The $x$, $y$ and $z$ coordinates of a single point are therefore denoted via $x_i$, $y_i$ and $z_i$.
Therefore, the point cloud $G$ is the discretization of the continuous geometry $\mathcal{G}$, and is stored as a $\mathbb{R}^{n\times3}$ matrix. 

Each geometry $\mathcal{G}$ used in this work,
was created via a set of $k \in \mathbb{N}$ generating geometry parameters $\mathcal{P} = \lbrace p_1, \ldots , p_k \rbrace$, $k = \lvert \mathcal{P}\rvert$
which can be stored as a vector $\vec{p} = (p_1, \ldots, p_k)\tp\in \mathbb{R}^{k}$.
We declare a geometry class as a set of geometries that share the same set of generating parameters, but with different values. 
For example, the circles with the center at the origin only need the radius $r$ as a geometry parameter to be constructed.
Thus, $k = 1$, $p_1 = r$ and $\mathcal{P} = \lbrace r\rbrace$ for this example.
Thinking of a rectangular cuboid, the number of geometry parameters would be three ($k = 3$), namely the width $p_1 = a$, the length $p_2 = b$, and the height $p_3 = c$. The parameter set then can be written as $\mathcal{P} = \lbrace a,b,c \rbrace$.

Generating a geometry means realizing the mapping from the generating parameter set $\mathcal{P}$ to the geometry $\mathcal{G}$, $f\colon \mathcal{P} \to \mathcal{G}$.
The inverse function $f^{-1}$ would then map the geometries back to the generating parameters, but in general the existence and realization of such a function is unknown. 

\subsection{Generating Random Vectors from Geometries}
As explained, we are dealing with 3-D point clouds $G \subset \mathbb{R}^3$, which can be represented with a matrix $G \in \mathbb{R}^{n \times 3}$.
Later on we see, that the PCA expects random vectors $\vec{x}$ as an input.
In the context of this work, each varied geometry is viewed as a realization of an underlying meta-geometry. 
Therefore, each vector created from a geometry is a random realization of an underlying distribution.
A simple approach to mapping a matrix to a vector is to put the columns below each other.
This process can be realized via the function $\vecop\colon \mathbb{R}^{n \times d} \to \mathbb{R}^{d\cdot n}$ and $\vecop^{-1}\colon \mathbb{R}^{d\cdot n} \to \mathbb{R}^{n \times d}$.
In the example of the point cloud $G$, this looks as follows:
\begin{equation}
    \vecop(G) = \vec{x} = \left( x_1, \ldots, x_n, y_1, \ldots, y_n, z_1, \ldots, z_n  \right)\tp \in \mathbb{R}^{3n} \point
\end{equation}
The inverse operation would look like
\begin{equation}
    \vecop^{-1}(\vec{x}) = G = 
    \begin{pmatrix}
        x_1 & y_1 & z_1 \\
        \vdots & \vdots & \vdots \\
        x_n & y_n & z_n
    \end{pmatrix} \in \mathbb{R}^{n \times 3} \point
\end{equation}
Thus, this function transforms each point cloud matrix $G_i$ into a vector $\vec{x}_i$, $i \in \lbrace 1,2, \ldots, m \rbrace$, and vice versa.
All geometries of a class are transformed into a vector and form a matrix $X$, that will be the input for the PCA.


\subsection{Standard Principal Component Analysis}
The PCA is a well-established statistical technique for dimensional reduction while preserving potentially most of the contained information.
It will map a dataset within a high-dimensional space to a dataset in a lower-dimensional space, or latent space, spanned by the orthogonal modes of variation of the data.
These modes are realized as vectors.

Now, we want to recall briefly the PCA method.
We start discussion with a dataset consisting of $m\in \mathbb{N}$ random vectors $\vec{x} \in \mathbb{R}^{l}$ of length $l \in \mathbb{N}$. 
In practice, we can assume that $m \gg l$.
These random vectors then will be merged into a matrix 
\begin{equation}
	X = \left[ \vec{x}_1, \vec{x}_2, \ldots ,\vec{x}_m \right]\in \mathbb{R}^{l\times m}
\end{equation}
that builds the base for further computations.

The first processing step is the centering of the data $X$.
Therefore, the mean of each row is calculated and stored in the vector $\vec{\mu}\in \mathbb{R}^{l}$.
Then this vector $\vec{\mu}$ will be subtracted from each random vector $\vec{x}$ in $X$, 
\begin{equation}
    \label{eq:pca_mean_substraction}
    X-\vec{\mu}\vec{1}\tp =: X_\mu \comma
\end{equation}
with $X_\mu$ the matrix of the centered data and $\vec{1}$ the all-ones vector of size $l$.

The second step is the computation of the sample covariance matrix 
\begin{equation}  \label{eq:cov_pca}
	C = \frac{1}{m-1} X_\mu X_\mu\tp \in \mathbb{R}^{l \times l} \point
\end{equation}

The third step is the computation of spectral decomposition of the covariance matrix $C$,
\begin{equation}
	C = V \Lambda V\tp \comma
\end{equation}
where $\Lambda = \diag(\lambda_1, \ldots, \lambda_l) \in \mathbb{R}^{l \times l}$ is the diagonal matrix with the sorted eigenvalues $\lambda_1 > \lambda_2 > \ldots >\lambda_l \geq 0$ of the matrix $C$.
The eigenvectors $\vec{v}_i$ corresponding to $\lambda_i$, are normalized $\lVert \vec{v}_i \rVert = 1$ and stored in the matrix $V = \left[\vec{v}_1, \ldots, \vec{v}_l \right]\in \mathbb{R}^{l\times l}$.
Since the matrix $C$ is symmetric by construction, the matrix $V$ will be an orthogonal matrix $\langle \vec{v}_i \mid \vec{v}_j \rangle = \delta_{ij}$, with Kronecker delta $\delta_{ij}$.

The final step, would be the computation of the principal components $Y$ of a random vector $\vec{x}$:
\begin{equation}
	y_i = \vec{v}_i\tp (\vec{x}-\vec{\mu}) = \langle \vec{v}_i \mid \vec{x} - \vec{\mu}\rangle \quad\text{or}\quad \vec{y} = V\tp (\vec{x}-\vec{\mu}) = \sum_{i = 1}^{l} \vec{e}_i\langle \vec{v}_i \mid \vec{x} - \vec{\mu}\rangle
\end{equation}
with $y_i$ denoting the $i$th principal component, $\vec{e}_i$ the $i$th unit vector, and $\langle \cdot \mid \cdot \rangle$ is the inner product.

The orthogonal modes of variation are thus defined by the eigenvectors, 
and the variation is encapsulated in the eigenvalues via $\var(y_i) = \lambda_i$.

Let us remark explicitly on some elementary PCA properties.
The eigenvalues and eigenvectors are linked to each other, and the eigenvalues are ordered. 
Therefore, the eigenvector belonging to the first eigenvalue, hereinafter referred to as the \emph{first eigenvector}, is the direction of the highest variance.
Thus, this axis stores the most information of the data put into the PCA. 
Conversely, the eigenvectors linked to eigenvalues close to zero store less useful information. 
The information along these axes is mostly stored in the mean vector.
The efficacy of PCA in reducing dimensionality is predicated on these considerations.

\subsubsection{Setup for Parameter Estimation\label{sec:parameter_estimation}}
Since a geometry can be reconstructed by a linear combination of the eigenvectors $V$, we assume a similar behavior for the generating parameters. 
These parameters would be a linear combination of some 
vectors $\vec{h}$ related to the parameters.
And these vectors $\vec{h}$, packed in the matrix $H$, are linearly combined in the same way as the geometry:
\begin{equation}
    \label{eq:connection_eq_our_approach}
    X_\mu = X - \vec{\mu}_X \vec{1}\tp = V Z \qquad \Longleftrightarrow \qquad P_\mu = P - \vec{\mu}_P\vec{1}\tp = H Z
\end{equation}
with the matrix $Z$ that produces the linear combination. 
A distinction between the mean vector of the points $\vec{\mu}_X$ and the mean vector of the parameter $\vec{\mu}_P$ is done by the indices $X$ and $P$. 

To determine the matrix $H$, the left equation of \eqref{eq:connection_eq_our_approach} is solved for the matrix $Z$.
Then, the matrix $Z$ is used to solve the right equation of \eqref{eq:connection_eq_our_approach} for $H$.
The left equation will be solved by multiplying $V\tp$ from the left: 
\begin{equation}
    V\tp X_\mu = V\tp V Z = Z \point
\end{equation}
Because the matrix $V$ is orthogonal, we have $V\tp = V^{-1}$.
The matrix $Z$ can then be put into the right equation of \eqref{eq:connection_eq_our_approach}: 
\begin{equation}
    \label{eq:paremeter_equation}
    P_\mu = H Z \quad \Longleftrightarrow \quad P_\mu = H V\tp X_\mu
\end{equation}
By use of the Moore-Penrose inverse \cite{Bjerhammar1951,Moore1920,Penrose1955}, or pseudoinverse, we can solve this regarding to $H$: 
\begin{equation}
    \label{eq:h_from_parameter_estimation}
    \begin{aligned}
		P_\mu & = H V\tp X_\mu \\
		P_\mu X_\mu\tp & = H V\tp X_\mu X_\mu\tp \\
		P_\mu X_\mu\tp \left(X_\mu X_\mu\tp \right)^{-1} & = H V\tp \\
		P_\mu X_\mu\tp \left(X_\mu X_\mu\tp \right)^{-1}V & = H 
	\end{aligned} 
\end{equation}
After computing the matrices $H$ and $Z$ the parameter vector $\vec{p}$, for a given single geometry $\vec{x}$, can be computed following \eqref{eq:paremeter_equation} via 
\begin{equation}\label{eq:parameter_estimation_our_approach}
    \vec{p} - \vec{\mu}_P = HV\tp (\vec{x} - \vec{\mu}_X) 
\end{equation}

\subsection{Joint PCA \label{sec:joint_pca}}
After discussing the standard PCA and a simple approach to estimate parameters using the eigenvectors, we want to discuss  the extended PCA, introduced in \cite{Serani2023}, that combines both. 
In the following, we will call it here \emph{joint PCA}.

The starting point is once again the matrix $X = \left[\vec{x}_1, \ldots, \vec{x}_m\right] \in \mathbb{R}^{l \times m}$.
However, the authors make use of two other matrices, $M$ a mass matrix and $W$ a weight matrix. 
The matrix $M = \diag(M_1, M_2, M_3) \in \mathbb{R}^{l\times l} $ contains the diagonal matrices $M_i = \diag(\Delta \mathcal{M}_1, \ldots, \Delta \mathcal{M}_\frac{l}{3}) \in \mathbb{R}^{\frac{l}{3}\times \frac{l}{3}}$ with the scalar values $\Delta \mathcal{M}_j$ for the $j$th point, meaning
\begin{equation}
	M = 
	\begin{pmatrix}
		M_1 & 0 & 0 \\
		0 & M_2 & 0 \\
		0 & 0 & M_3 \\
	\end{pmatrix} 
	\quad \text{with} \quad 
	M_i = \begin{pmatrix}
		\Delta \mathcal{M}_1 & 0 & \ldots & 0 \\
		0 & \Delta \mathcal{M}_2 & \ddots & \vdots \\
		\vdots & \ddots & \ddots & 0 \\
		0 & \ldots & 0 & \Delta \mathcal{M}_\frac{l}{3}
	\end{pmatrix} \point
\end{equation}
The other matrix $W \in \mathbb{R}^{l \times l}$ is defined as 
\begin{equation}
	W =
	\begin{pmatrix}
		W_x & 0 & 0 \\
		0 & W_y & 0 \\
		0 & 0 & W_z \\
	\end{pmatrix} 
	\quad \text{with} \quad  
	W_i = \begin{pmatrix}
		\rho_1 & 0 & \ldots & 0 \\
		0 & \rho_2 & \ddots & \vdots \\
		\vdots & \ddots & \ddots & 0 \\
		0 & \ldots & 0 & \rho_\frac{l}{3}
	\end{pmatrix}\in \mathbb{R}^{\frac{l}{3}\times \frac{l}{3}} 
    \point
\end{equation} 
We will notice later on, that the random vectors will have a length $l$ of three times the number of points in the geometry. 
Hence, a condition that ensures $\frac{l}{3} \in \mathbb{N}$ can be omitted. 
This is also indicated by the indices $x$, $y$, and $z$. 

The eigenvalue problem proposed in \cite{Serani2023} reads as 
\begin{equation}\label{eq:joint_pca_EV_problem}
	\underbrace{\frac{1}{m} XX\tp MW}_{:=C} V = CV =  V \Lambda 
    \quad \Longleftrightarrow \quad
    XX\tp MW = m V\Lambda V\tp
\end{equation}
Since there is no explicit mention in \cite{Serani2023} that the data $X$ are centered, we assume this for further argumentation.   

After recalling the basic PCA setting, the extension to include the parameters is done in \cite{Serani2023} via merging
\begin{equation}
	\vec{x}^{\,\mathrm{L}} = \left( x_1, \ldots , x_l , p_1, \ldots, p_k \right)\tp = \left(\vec{x}, \vec{p}\right)\tp \in \mathbb{R}^{l+k}
\end{equation}
the points and the parameters.
With this enlarged random vector, we can construct an enlarged random matrix $X^{\mathrm{L}} = \left[ \vec{x}^{\,\mathrm{L}}_{1}, \ldots , \vec{x}^{\,\mathrm{L}}_{m} \right] \in \mathbb{R}^{(l+k) \times m}$.

Since the matrix $X^{\mathrm{L}}$ has a different size than the matrix $X$, the matrices $M$ and $W$ needed to be enlarged as well. 
The matrices $M^{\mathrm{L}},W^{\mathrm{L}}  \in \mathbb{R}^{(l+k) \times (l+k)}$ will be
\begin{equation}
	M^{\mathrm{L}} = 
	\begin{pmatrix}
		M & 0 \\
		0 & I_k \\
	\end{pmatrix} 
    \quad \text{and} \quad
    W^{\mathrm{L}} = 
	\begin{pmatrix}
		W & 0 \\
		0 & 0 \\
	\end{pmatrix}
\end{equation}
with $I_k \in \mathbb{R}^{k \times k}$ the identity matrix.
The eigenvalue problem \eqref{eq:joint_pca_EV_problem} transforms to its final extended PCA version 
\begin{equation}\label{eq:joint_pca_EV_problem_L}
	\underbrace{\frac{1}{m} X^{\mathrm{L}} \left(X^{\mathrm{L}}\right)\tp M^{\mathrm{L}} W^{\mathrm{L}}}_{:=C^{\mathrm{L}}} V^{\mathrm{L}} = C^{\mathrm{L}} V^{\mathrm{L}} =  V^{\mathrm{L}} \Lambda^{\mathrm{L}}
\end{equation}
with $V^{\mathrm{L}}  \in \mathbb{R}^{(l+k) \times (l+k)}$.

Next, the authors of \cite{Serani2023} provide us with a discussion of the relation between $V$ and $V^\mathrm{L}$. 
The enlarged random matrix  $X^\mathrm{L}$ can be written using the random matrix $X$ and a parameter matrix $P$ via 
\begin{equation}
	X^{\mathrm{L}} = \begin{pmatrix} 
	X \\ P \end{pmatrix} \in \mathbb{R}^{(l+k) \times m} \point
\end{equation} 
Like the construction of the matrix $C$, 
the enlarged matrix $C^\mathrm{L}\in \mathbb{R}^{(l+k) \times (l+k)}$
can be obtained via the following calculations: 
\begin{equation}
\label{eq:CL_transformations}
	\begin{aligned}
		C^{\mathrm{L}} &= \frac{1}{m} X^{\mathrm{L}} \left(X^{\mathrm{L}}\right)\tp M^{\mathrm{L}} W^{\mathrm{L}} =  \frac{1}{m} \begin{pmatrix} X \\ P \end{pmatrix} 
		\begin{pmatrix} X\tp & P\tp \end{pmatrix}
		\begin{pmatrix}
			M & 0 \\
			0 & I \\
		\end{pmatrix}
		\begin{pmatrix}
			W & 0 \\
			0 & 0 \\
		\end{pmatrix} \\
		&= \frac{1}{m} \begin{pmatrix} X \\ P \end{pmatrix} 
		\begin{pmatrix} X\tp & P\tp \end{pmatrix}
		\begin{pmatrix}
			MW& 0 \\
			0 & 0 \\
		\end{pmatrix} \\
		&= \frac{1}{m} \begin{pmatrix} X \\ P \end{pmatrix}
		\begin{pmatrix} X\tp MW & ~~ 0 \end{pmatrix} = \frac{1}{m} 
		\begin{pmatrix}
			X X\tp MW & ~~0 \\
			P X\tp MW & ~~0
		\end{pmatrix} = \begin{pmatrix}
			C & 0 \\
			C_P & 0
		\end{pmatrix} 
		\end{aligned}
\end{equation}
From these, we notice that $C^\mathrm{L}$ makes use of the matrix $C$ from the pre-extended version and a new matrix  $C_P \in \mathbb{R}^{k \times l}$. 
Now, the eigenvalues of the matrix $C^{\mathrm{L}}$ can be computed by the Laplace expansion as 
\begin{equation}
	\det\left(C^{\mathrm{L}} - \lambda I\right) = \begin{vmatrix}
		C - \lambda I_{l} & 0 \\
		C_P & -\lambda I_k
	\end{vmatrix} = \det(C - \lambda I_{l}) \cdot \det(-\lambda I_k) = 0 \point
\end{equation}
Having a suitable information-carrying eigenvalue $\lambda\neq0$, 
$\lambda$ needs to satisfy $\det\left(C^{\mathrm{L}} - \lambda I\right) = 0$.
Since $\lambda$ is non-zero, we have $\det(-\lambda I_k) \neq 0$ and therefore $\lambda$ need to satisfy $\det(C - \lambda I_{l}) = 0$ too. 
This proves that the non-zero eigenvalues of $C^{\mathrm{L}}$ and $C$ are identical.
Therefore, the matrix $\Lambda^{\mathrm{L}}$ can be constructed with the eigenvalue matrix $\Lambda$ like
\begin{equation}
	\Lambda^{\mathrm{L}} = \begin{pmatrix}
		\Lambda & 0 \\
		0 & 0
	\end{pmatrix} \in \mathbb{R}^{(l+k) \times (l+k)} \point
\end{equation}
Now, $V^\mathrm{L}$ can be constructed with the four matrices $V_1\in \mathbb{R}^{l \times l}$, $V_2\in \mathbb{R}^{l \times k}$, $V_3\in \mathbb{R}^{k \times l}$ and $V_4\in \mathbb{R}^{k \times k}$ as 
\begin{equation}
	V^{\mathrm{L}} = \begin{pmatrix}
		V_1 & V_2 \\
		V_3 & V_4
	\end{pmatrix} \in \mathbb{R}^{(l+k) \times (l+k)}
\end{equation}
and then the eigenvalue problem \eqref{eq:joint_pca_EV_problem_L} can be rewritten as
\begin{equation}
    \label{eq:joint_pca_ev_evaluation}
    \begin{pmatrix}
        C & 0 \\
        C_P & 0 
    \end{pmatrix}
    \begin{pmatrix}
        V_1 & V_2 \\
        V_3 & V_4 
    \end{pmatrix} = 
    \begin{pmatrix}
        V_1 & V_2 \\
        V_3 & V_4 
    \end{pmatrix}
    \begin{pmatrix}
        \Lambda & 0 \\
        0 & 0 
    \end{pmatrix} \Longleftrightarrow
    \begin{pmatrix}
        CV_1 & CV_2 \\
        C_P V_1 & C_P V_2 
    \end{pmatrix} = 
    \begin{pmatrix}
        V_1\Lambda & 0 \\
        V_3\Lambda & 0 
    \end{pmatrix} \point
\end{equation}
A comparison of the coefficients, yielding that $V_1 = V$ to be in line with \eqref{eq:joint_pca_EV_problem}. 
For the matrix $V_2$ we have $V_2 = 0$, and since the matrix $V_4$ was not used, it can be set to $V_4 = 0$.
The last unknown matrix is $V_3$, which can be computed with the help of the expression for $C_P$ obtained from \eqref{eq:CL_transformations}: 
\begin{equation}
    \label{eq:parameter_eigenvector_construction}
	\begin{aligned}
		C_P V  = V_3 \Lambda 
        ~ \Longleftrightarrow ~
		\frac{1}{m} P X\tp MW V  = V_3 \Lambda 
        ~ \Longleftrightarrow ~
		\frac{1}{m} P X\tp MW V \Lambda^{-1} = V_3 ~.
	\end{aligned}
\end{equation}
The matrix $V_3$ stores the vectors related to the parameters, which can be seen later on when $V^\mathrm{L}$ is used to reconstruct an element $\vec{x}^{\,\mathrm{L}}$.
Keeping the notation similar to the standard PCA parameter estimation, the matrix $V_3$ is already renamed here in $V_3 = H$. 
From these calculations, conclusions about the eigenvectors can be made. 
In total, we have after \eqref{eq:joint_pca_ev_evaluation}  
\begin{equation}
    V^\mathrm{L} = \begin{pmatrix}
			V & 0 \\
			H & 0 
		\end{pmatrix} 
        = \left[ \vec{v}^{\,\mathrm{L}}_1, \ldots, \vec{v}^{\,\mathrm{L}}_l, \vec{0}, \ldots ,\vec{0} \right]  
\end{equation}
with 
\begin{equation}
    \vec{v}^{\,\mathrm{L}}_i = \left(\vec{v}_i, \vec{h}_i\right)\tp = \left( v_1, \ldots, v_{l}, h_1, \ldots h_k \right)\tp \point
\end{equation}

This discussion showed, that because of the specific choice of $M^{\mathrm{L}}$ and $W^{\mathrm{L}}$, merging the parameter vectors $P$ with random vectors $X$ will not change eigenvalues and eigenvectors.

\subsubsection*{Setup for Parameter Estimation}

After the matrix $V^{\mathrm{L}}$ is learned, the parameter estimation for a geometry $\vec{x}$ can be realized with 
\begin{equation}
\label{eq:estimation_jpca_transfo}
    \vec{x}^{\,\mathrm{L}} = V^{\mathrm{L}} \vec{z}^{\,\mathrm{L}} 
    \quad \Longleftrightarrow \quad
    \begin{pmatrix}
        \vec{x} \\ \vec{p} 
    \end{pmatrix} = 
    \begin{pmatrix}
        V & 0 \\ H & 0 
    \end{pmatrix} 
    \begin{pmatrix}
        \vec{z}_1 \\ \vec{z}_2 
    \end{pmatrix} =
    \begin{pmatrix}
        V \vec{z}_1 + 0 \vec{z}_2 \\ H \vec{z}_1 + 0\vec{z}_2 
    \end{pmatrix} =
    \begin{pmatrix}
        V \vec{z}_1 \\ H \vec{z}_1
    \end{pmatrix}~.
\end{equation}
Comparing these results to \eqref{eq:connection_eq_our_approach} it can be seen that both approaches for estimation are identical. 
The difference is, that the parameter vectors $H$ do not need to be computed separately. 

To obtain the parameter vector $\vec{p}$ from \eqref{eq:estimation_jpca_transfo}, we need to decouple it and get
\begin{equation}
    \label{eq:connection_eq_serani}
    \vec{x} = V \vec{z}_1 
    \quad \mathrm{and} \quad
    \vec{p} = H\vec{z}_1 \point
\end{equation}
Estimating the parameters for a given random vector $\vec{x}$ can then be done via
\begin{equation}\label{eq:parameter_estimation_serani}
    \vec{p} = H V\tp \vec{x} \quad \mathrm{with} \quad \vec{y}_1 = V\tp \vec{x} \point
\end{equation}


\subsection{Relation between Joint PCA and Standard PCA}

Because the standard PCA offers a more interpretable approach, we want to compare both ways of parameter estimation.
First, we notice that the equations, \eqref{eq:parameter_estimation_our_approach} and \eqref{eq:parameter_estimation_serani}, look similar due to notation.
Therefore, equate the estimation results of both approaches is the start for the investigation.

For the further transformations, we want to indicate with an index $\mathrm{S}$ the results from the paper of Serani \cite{Serani2023}, and with an index $\mathrm{O}$ the results from our approach.
Additionally, we omit the $\mu$ notation from our approach, and declare that the matrix $X$ has zero mean.
We start with the left equations of \eqref{eq:parameter_estimation_serani} and \eqref{eq:parameter_estimation_our_approach}. 
Since we want to know the parameter for the same random vector $\vec{x}$, this vector can be omitted, i.e. 

\begin{equation}
    H_{\mathrm{S}} V_{\mathrm{S}}\tp \vec{x} = H_{\mathrm{O}} V_{\mathrm{O}}\tp \vec{x} \quad \Longleftrightarrow \quad  
    H_{\mathrm{S}} V_{\mathrm{S}}\tp = H_{\mathrm{O}} V_{\mathrm{O}}\tp    \point
\end{equation}
On the left side, we can substitute the matrix $H_{\mathrm{S}}$ using its definition; compare \eqref{eq:parameter_eigenvector_construction}.
The same is done with $H_{\mathrm{O}}$ and \eqref{eq:h_from_parameter_estimation}, penultimate equation, yielding 
\begin{equation}
    \label{eq:compare_step_2}
    \frac{1}{m} PX\tp MW V_{\mathrm{S}} \Lambda_{\mathrm{S}}^{-1}V_{\mathrm{S}}\tp 
        = PX\tp\left(XX\tp\right)^{-1} \point
\end{equation}
Reminder, the matrices $X$ and $P$ store the points and parameters of the geometries and are the same for both approaches. 
And since the mass and weight matrix $M$ or $W$ only appear in the joint PCA method, an index is not needed. 

On the left side of this equation, there is the part of 
\begin{equation*}
    \frac{1}{m}V_{\mathrm{S}} \Lambda_{\mathrm{S}}^{-1}V_{\mathrm{S}}\tp = \left(mV_{\mathrm{S}} \Lambda_{\mathrm{S}}V_{\mathrm{S}}\tp\right)^{-1} \point
\end{equation*}
Comparing this equation to \eqref{eq:joint_pca_EV_problem} it rewrites to 
\begin{equation*}
    \left(mV_{\mathrm{S}} \Lambda_{\mathrm{S}}V_{\mathrm{S}}\tp\right)^{-1} = \left(XX\tp MW\right)^{-1} \point
\end{equation*}
Putting this into \eqref{eq:compare_step_2}, we get 
\begin{equation}
    PX\tp MW \left(XX\tp MW\right)^{-1} = PX\tp\left(XX\tp\right)^{-1} \point
\end{equation}
The product $PX\tp$ can be omitted in both equations, and moving $M$ and $W$ out of the inverse, we get 
\begin{equation}
    MW W^{-1}M^{-1}\left(XX\tp \right)^{-1} = \left(XX\tp\right)^{-1}\point 
\end{equation}

The first part on the joint PCA side of this equation is $MW W^{-1}M^{-1} = I$, and the overall relation evolves to $XX\tp = XX\tp$, which is the starting point of the PCA in both cases. 

Hence, we have shown that the effect of $M$ and $W$ gets canceled out within the process of parameter  estimation. 
Since both methods thus lead to the same parameter, we will omit the joint PCA method in the following considerations. 
Following the standard PCA is used because it provides a simpler framework for data exploration.


\subsection{Quality of the PCA\label{sec:r_test}}

In practice, a tool to evaluate the quality of the PCA is needed. 
Within the paper, the cumulative ratio of total variation (CRV) is chosen \cite[Chap. 6.1.1]{Jolliffe2002}.  
It computes the ratio of the sum of the first $t$ eigenvalues to the sum of all eigenvalues as  
\begin{equation}
    \label{eq:crv}
	\mathrm{CRV}_t = \dfrac{\sum_{i=1}^{t} \lambda_i}{\sum_{i=1}^{l} \lambda_i} \in [0,1] \point
\end{equation}
Ratios close to 1 for $t\ll l$ indicate that almost all information is stored in the first $t$ eigenvalues.
In this paper, the index $t$ for which a given ratio $\mathrm{CRV}_t$ is reached is of particular importance. 


\section{Experiments \label{sec:Experiments}}

Within this section, we explain our test geometry classes and the corresponding generating parameters. 
We move on to the visual exploration of the mean and eigenvectors produced by the PCA in the context of these geometries as input. 
We finalize the investigation with the results of the parameter estimation regarding the results of the visual explorations. 

\subsection{Dataset Geometries}

\begin{figure}[t]
    \centering
    \includegraphics[width=0.475\linewidth]{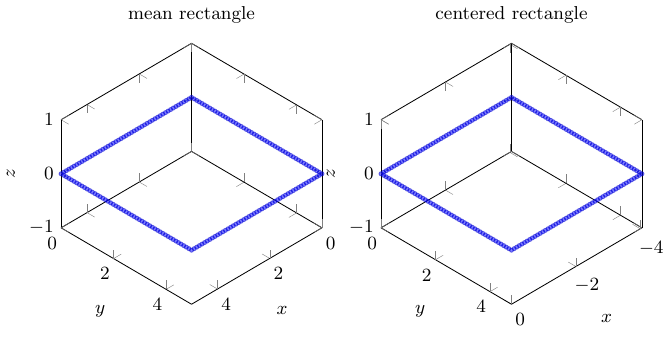}
    \includegraphics[width=0.475\linewidth]{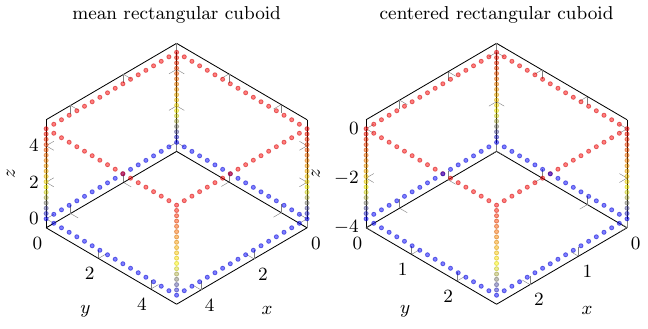}
    \caption{The rectangle and rectangular cuboid geometry classes. In the first and third image from the left, the mean geometries are plotted. In the second and fourth image, there are the corresponding centered geometries. For both classes, there is no change in geometry class. We have a good interpretable knowledge representation for both classes.}
    \label{fig:rectangle_mean_and_wo_mean}
\end{figure}

The dataset consists of various geometry classes.
Every geometry class has $2\,000$ variations of a given geometry as members. 
The classes considered are a \emph{rectangle}, as a simple 2-D example, with a random length $a$ and width $b$, both with a range of $[0,10)$.
The parameter set is $\mathcal{P}_{\mathrm{r}} = \lbrace a,b \rbrace$.
The visualization of the rectangles can be seen in Figure \ref{fig:rectangle_mean_and_wo_mean}, the left two images. 

We added a \emph{rectangular cuboid} class as a leap into 3-D examples, with random length $a$, width $b$, and height $c$, again in a range of $[0,10)$. 
The parameter set for this geometry would be $\mathcal{P}_{\mathrm{c}} = \lbrace a,b,c \rbrace$.
An example of this geometry class can be seen in Figure \ref{fig:rectangle_mean_and_wo_mean}, the right two images.

The next class in the dataset is a \emph{helix} geometry, which can be constructed via the following equation:
\begin{equation}
    \label{eq:helix_creation}
    \vec{f}(\mathcal{P}_{\mathrm{h}}) = (r \cos(2\pi n), r \sin(2\pi n), h)\tp \point
\end{equation}
The parameter set consists of the radius $r$, the height $h$, and the number of turns $n$ in the helix, so that $\mathcal{P}_{\mathrm{h}} = \lbrace r, h, n \rbrace$. 
Additionally, we use a simplified creation where the number of turns is fixed, i.e. $\mathcal{P}_{\mathrm{hs}} = \lbrace r, h \rbrace$. 
The range for the radius and height is $[0,10)$, and the range for the number of turns is $[0,5)$. 
For the simplified helix, the number of turns is fixed at 5 turns. 
A visualization of the two helix classes can be seen in Figure \ref{fig:simple_helix_mean_and_wo_mean}.

If not mentioned otherwise, the number of points within a point cloud is set to 200 points.  
Therefore, the random vectors created from these geometries have a dimension of $\mathbb{R}^{3\cdot200}$. 
So the variable $l$, which was used to indicate the length of the random vector that is put into the PCA, is $600$. 

Together with these basic geometry classes, there are two more complex geometry classes.
These are a fan blade geometry and a tube with a round and rectangular end, cf. left and right two columns of Figure \ref{fig:fan_blade_mean_and_wo_mean}.
The parameter set of these classes consists of $\lvert \mathcal{P}_{\mathrm{fb}} \rvert = k = 12$ parameters for the fan blade class and $\lvert \mathcal{P}_{\mathrm{t}} \rvert = 14$ for the tube geometry.

\subsection{Preprocessing of the data}

\subsubsection{Train and test data}
Like mentioned, there are $2\,000$ variations within a geometry class. 
These were split in a ratio $9:1$ into a training set containing $1\,800$ random geometries per class and $200$ for testing the parameter estimation. 
Therefore, the parameter $m$, which was indicating the number of random vectors used in the PCA, is set to $1\,800$.  

\subsubsection{What are we actually learning using PCA?}
The common visualization of how PCA works, is having a cigar-shaped, or elliptical, point cloud of 2-D random vectors, somewhere in a Cartesian coordinate system. 
Subtracting the mean vector of the data will lead to a cigar-shaped point cloud at the origin of the axes. 
And the PCA will provide us with two eigenvectors, the one belonging to the higher eigenvalue in the longer direction and the other one in the shorter direction.

Although this picture is correct, it nevertheless may induce the idea that, when learning a 3-D geometry, one starts with random vectors $\in \mathbb{R}^3$, that are randomly distributed throughout the space, and this random distribution has the shape of the desired geometry.
With this kind of thinking, centering the geometry, means that the geometric center is moving to the coordinate center. 
But that is not what is happening here.
With the approach within this paper, each geometry with $n$ points is transformed to a random vector with a length of $3n$. 
Therefore, we end up with $m$ vectors of length $3n$ forming a point cloud in a $\mathbb{R}^{3n}$ space. 
This 3$n$-D point cloud is then centered.
Therefore, the mean vector $\vec{\mu}$ is an element of $\mathbb{R}^{3n}$ too, and can be transformed back to a 3-D geometry.
The study of the mean vector is part of the next section.


\subsection{Impact of the Mean}

\begin{figure}
    \centering
    \includegraphics[width=0.475\linewidth]{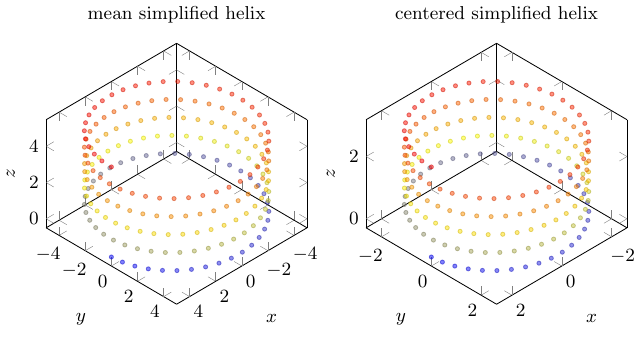}
    \includegraphics[width=0.475\linewidth]{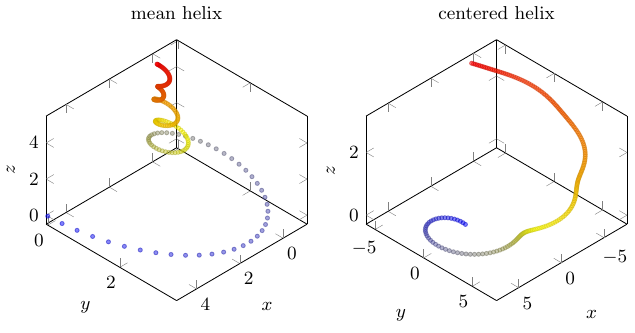}
    \caption{The simplified helix and helix geometry classes. In the first and third image from the left, the mean geometries are plotted. In the second and fourth image, there are the corresponding centered geometries. The simplified helix class does not change, but the helix class does. Therefore, the interpretability of the knowledge representation of the simplified helix is better than that of the helix.}
    \label{fig:simple_helix_mean_and_wo_mean}
\end{figure}

A crucial step within the PCA is the centering of the data; see Equation \eqref{eq:pca_mean_substraction}. 
During this step, the vectorized geometries get subtracted by the mean geometry vector. 
This process could alter the visual appearance of a geometry. 
Within this work, we will denote this with a change in geometry class. 
In these cases, where the change occurs, the PCA does not learn the main modes of variation of the desired geometry, but a different one.
And consequently, the knowledge representation gets harder to interpret. 

Within this visual investigation, the mean geometry is plotted on the left, and on the right, we provide the plot of an example of the geometry class reduced by the mean. 
Consequently, any alteration in the mean geometry and/or the subtracted geometry by centering is directly apparent. 
 
We start with showing cases where no change in class is happening. 
The mean of the rectangle geometry, ref. Figure \ref{fig:rectangle_mean_and_wo_mean} left image.
The second left image belongs to a rectangle that is already centered. 
Comparing the left image to the second left image, it is noticeable, that both geometries still belong to the rectangle geometry class.  
Since the value for the width and height of the rectangle was $<10$, the mean geometry still has a height and width of $\approx 5$.

Similar observations can be made for the rectangular cuboids in Figure \ref{fig:rectangle_mean_and_wo_mean}, right two images. 
Like in the 2-D scenario, the geometry class of the mean geometry and the subtracted geometry is unchanged.  

Next, we investigate the geometries of the simplified helix, where the number of turns is fixed, and the helix. 
In one of these geometries, the aforementioned change in geometry class will occur.   
The visualizations of the simplified helix and  helix are presented in Figure \ref{fig:simple_helix_mean_and_wo_mean}, left two and right two images.
We notice, that the simplified helix, does not change the geometry class, whereas the normal helix changes it. 
Both the mean geometry and the reduced geometry are no longer part of the helix class. 

\begin{figure}
    \centering
    \includegraphics[width=0.475\linewidth]{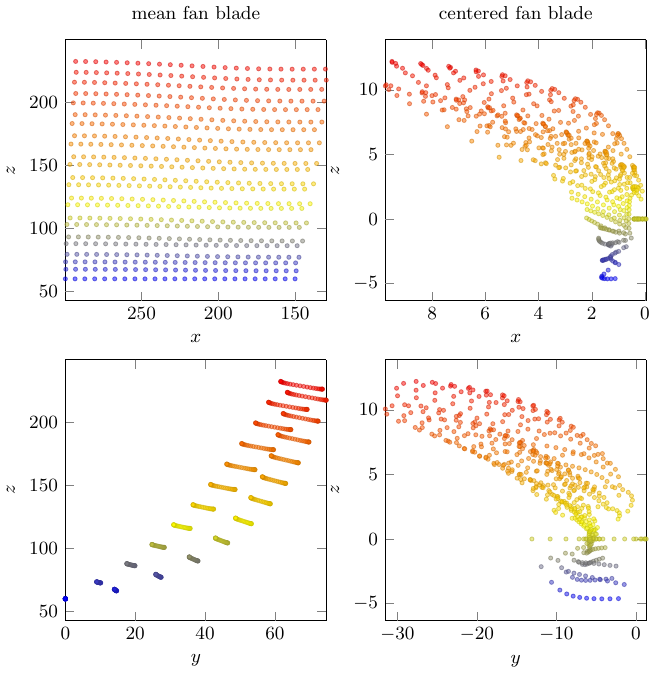}
    \includegraphics[width=0.475\linewidth]{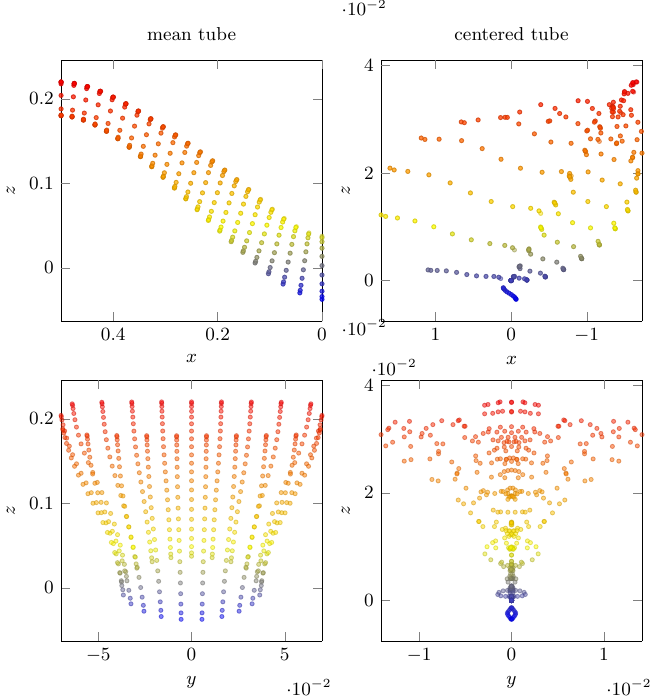}
    \caption{
        \label{fig:fan_blade_mean_and_wo_mean}
        The fan blade and tube geometry classes. In the first and third column from the left, the mean geometries are plotted from two different viewpoints. In the second and fourth column, there are the corresponding centered geometries, plotted again in two viewpoints. Noticeably, both geometry classes get changed during the centering process.
        Both geometries have a knowledge representation that is challenging to interpret. 
        }
\end{figure}

The images for the fan blade and tube geometry can be seen in Figure \ref{fig:fan_blade_mean_and_wo_mean}, left two columns and right two columns. 
Since we deal with more complex objects, we present two different viewpoints for each geometry, which are positioned above each other. 

In contrast to the helix geometry, both complex geometries keep their class, when computing the mean geometry. 
But, when contemplating the centered geometries, these objects no longer belong to the fan blade, respectively, tube class. 

\subsubsection*{Discussion:}

These visual investigations separate the used geometry classes into two parts. 
One part consists of the geometry classes that \emph{do not} change the class during centering, namely rectangle, rectangular cuboid, and simplified helix. 
The geometry classes that \emph{do} change the class, namely helix, fan blade, and tube, are in the second part. 

With the presented geometry classes, all cases are covered. 
The rectangle, rectangular cuboid, and simplified helix cover the case that the classes stay the same. 
The knowledge representation of these geometries is easy to interpret. 
With the complex geometries, the case that only the class of the reduced geometry changes is covered. 
And last, the helix covers the case that mean and reduced geometries have a class change. 
For these three geometries, the knowledge representation is difficult to interpret.
We note, that only in the first case, where no class change is happening, we do learn the intended geometry. 
In the other two cases, we learn a different geometry class. 

For further investigations, we want to separate the geometry classes into two sets.  
One, where the geometry classes were not changed, and one where they were. 
For convenience, the set of classes with no change will be called the \emph{first} set, and the other one will be the \emph{second} set; cf. Table \ref{tab:r_test_results}.

\subsection{Eigenvalues and Eigenvectors}

The learning results of the PCA consist of a set of eigenvalues and eigenvectors for each geometry class.
In the first part of the section, we look at the eigenvalues of all geometry classes, and in the second part, we present the selected eigenvectors of some geometry classes.

\subsubsection{Eigenvalues}

\begin{table}
    \centering
    \begin{tabular}{c|c|c|c|c||c}
        & Geometry class  & $\mathrm{CRV}_t = 0.9$ & $\mathrm{CRV}_t = 0.95$ & $\mathrm{CRV}_t = 0.99$ & $k$ \\ \hline
        \multirow{3}{*}{first set}& rectangle  & 2 & 2 & 2 & 2\\
        & rectangular cuboid  & 3 & 3 & 3 & 3 \\
        & simple helix  & 2 & 2 & 2 & 2\\ \hline
        \multirow{3}{*}{second set}& helix   & 10 & 11 & 12 & 3\\
        & fan blade  & 3 & 4 & 9 & 12\\
        & tube   & 2 & 3 & 6 & 14
    \end{tabular}
    \caption{\label{tab:r_test_results}
    Here, the number of eigenvalues $t$ needed to get a ratio of $0.9 \widehat{=} 90\%$, $0.95 \widehat{=} 95\%$, and $0.99 \widehat{=} 99\%$ for the geometry classes is presented. 
    The number of generating parameters $k$ is incorporated into the table for the purpose of comparison.
    }    
\end{table}

As a brief reminder, the eigenvalues of the PCA are in descending order, and along the eigenvector axis of higher eigenvalues, we have the most variance in the data.

To quantify how much information is stored in the eigenvalues, we make use of the CRV equation \eqref{eq:crv}.
We are interested in how many eigenvalues $t$ are needed to get $0.9 \widehat{=} 90\%$, $0.95 \widehat{=} 95\%$, and $0.99 \widehat{=} 99\%$ of the total information. 
The values $t$ for the different ratios can be seen in Table \ref{tab:r_test_results}.
The last column of this table, presents the number of generating parameters $k$ used to create the geometry. 

We expect, that $t$ fluctuates around the number of generating parameters $k$. 
This assumption is justified, because the generating parameters themselves store geometry-specific properties.   

For the geometry classes within the first set, the number of eigenvalues $t$ keeps constant over the chosen values $\mathrm{CRV}_t$.
For the rectangle and the simplified helix, the value is $t=2$. 
The rectangular cuboid stores all his information in $t =3$ eigenvalues.  
Additionally, the value $t$ matches the number of generating parameters $k$.  

For the geometry classes in the second set, we notice a different behavior. 
First, the value $t$ does not keep constant for different $\mathrm{CRV}_t$ values. 
Second, the values do not approach the number of generating parameters $k$.
Especially, the fan blade and tube geometry class has a significant jump in the numbers, when going from a ratio of $0.95$ to $0.99$. 
The third observation is, that the fan blade and tube geometry class needs fewer eigenvalues than generating parameters to store a meaningful amount of information, and the helix needs more. 

The results from the first geometry class set indicate a relation between $t$ and $k$. 
This relation cannot be seen for the classes within the second set. 

Detailed plots of the eigenvalue distributions for the geometry classes can be seen in the appendix Figure \ref{fig:eig_val_all_geometries}.

\subsubsection{Eigenvectors}

After the discussion about how many eigenvectors are needed to cover most of the information of the geometries, the next objects of study are the eigenvectors related to the meaningful eigenvalues. 
Since the eigenvectors can be transformed back to geometries, we like to employ the name eigengeometries as a reference to the eigenvectors. 
A crucial step in parameter estimation is to find a mixing matrix $Z$ that finds a linear combination of the eigengeometries to recover a certain geometry; see left Equations \eqref{eq:connection_eq_our_approach} and \eqref{eq:connection_eq_serani}.
With this in mind, the assumption is that the eigenvectors share basic characteristics of their corresponding geometry class.

\begin{figure}[t]
	\centering
    \includegraphics[width = \linewidth]{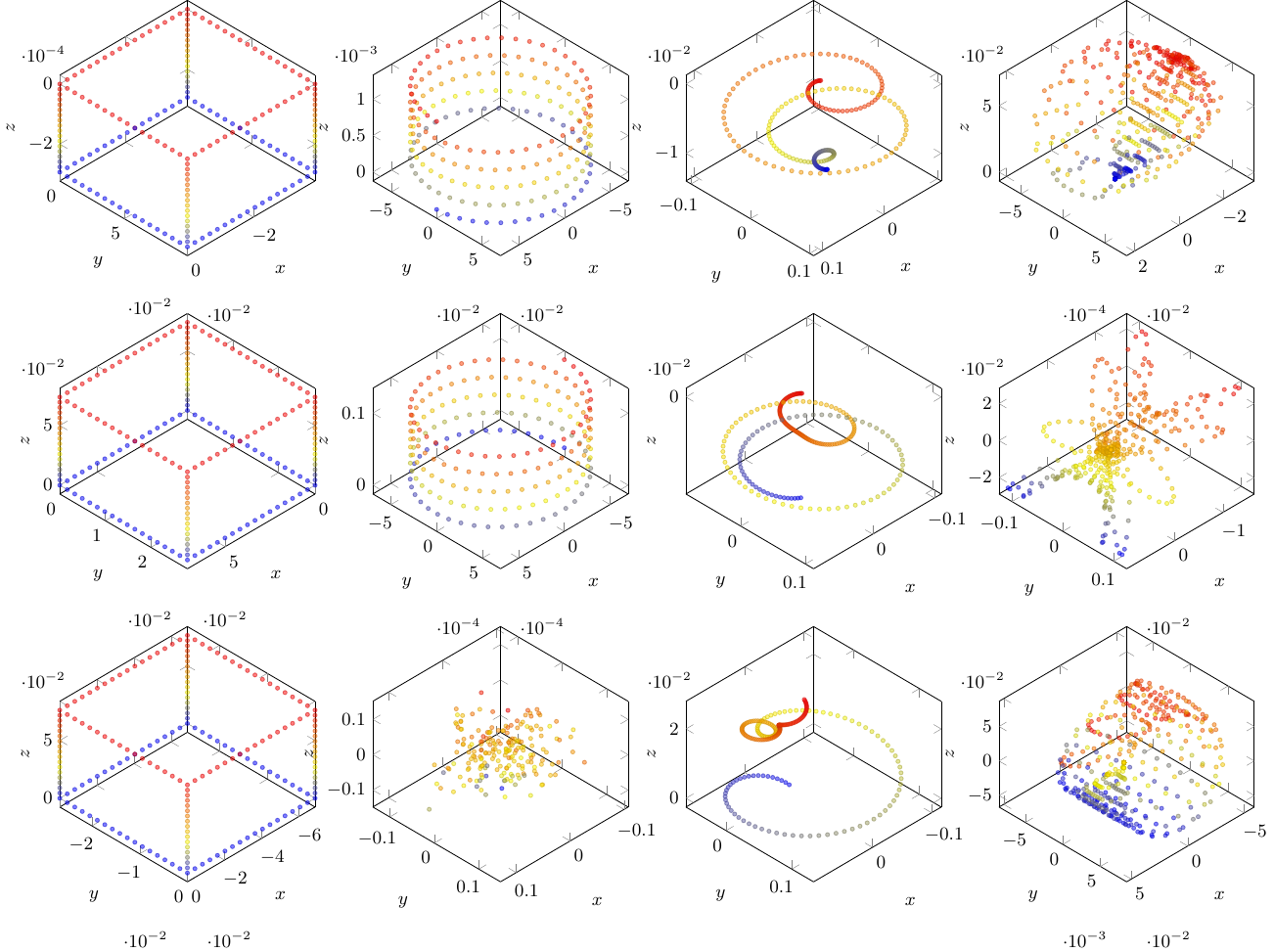}
	\caption{\label{fig:various_eig_vecs}
        The first three eigenvectors of selected geometry classes from the two sets are visualized. 
        In the first two columns, the classes rectangular cuboid and simplified helix, as representatives of the first set, are presented. 
        In the last two columns, the helix and tube geometries from the second set are presented. 
        The first row shows the first eigenvectors, the second row the second ones, and in the third row the third eigenvectors can be seen, respectively. 
        }
\end{figure}

In Figure \ref{fig:various_eig_vecs} we present row-wise the first three eigengeometries.
From the left to the right column, the eigengeometries of the rectangular cuboid, simplified helix, helix, and tube classes are shown. 
Since the rectangle behaves like the rectangular cuboid and has the same amount of information-carrying eigenvalues as the simplified helix, this class is omitted.
The fan blade geometry is omitted because they will not provide additional insights compared to the tube geometry. 

It is evident that for geometries where the class was not changed during the centering, the eigenvectors also did not change the geometry class. 
This indicates, that the eigengeometries are directly linked to the geometry class after subtracting the mean vector. 

The first eigengeometry of the rectangle is compressed along the $z$-axis, seen by the factor $10^{-4}$ next to this axis.
In contrast to the factor of the $x$ and $y$ axis, which is $10^{-2}$ for all three eigengeometries.  
The second and third eigengeometries have almost the same scale in the $z$-direction, but opposite ranges along the $x$ and $y$ axis.
Therefore, it seems reasonable that, with these three eigengeometries, an arbitrarily centered rectangular cuboid can be constructed. 

Similar observations can be made for the simplified helix eigengeometries. 
The first eigengeometry regulates mostly the radius of a helix. 
With the second eigengeometry, the height will be controlled. 
For this class, it seems intuitive, that these two eigengeometries form a well-interpretable knowledge representation of the simplified helix geometry class.
Looking at the third eigengeometry, it has no similarities to a helix. 
Together with Table \ref{tab:r_test_results} it obvious that this third eigengeometry stores no necessary information. 

For the two geometry classes from the second set, it is more challenging to notice the shared geometric characteristics of the corresponding geometry class.
Remember, during the reconstruction, the linear combination of the eigengeometries creates a centered geometry. 
The mean geometry needs to be added to this centered geometry to fully create an element of the corresponding geometry class. 
This fact also applies to the classes from the first set, but since the geometry class is not changed in any stage of the PCA, this argument can be neglected. 
However, the eigengeometries of the helix and tube classes lack the visual explainability of the geometry classes in the first set. 

The visual inspection of the eigengeometries shows, that the geometry classes in the first set, share expected characteristics of the corresponding geometry class. 
For the classes of the second set, an intuitive knowledge representation is not possible. 

\subsection{Parameter Estimation}

\begin{figure}[t]
	\centering
    \includegraphics[width = \textwidth]{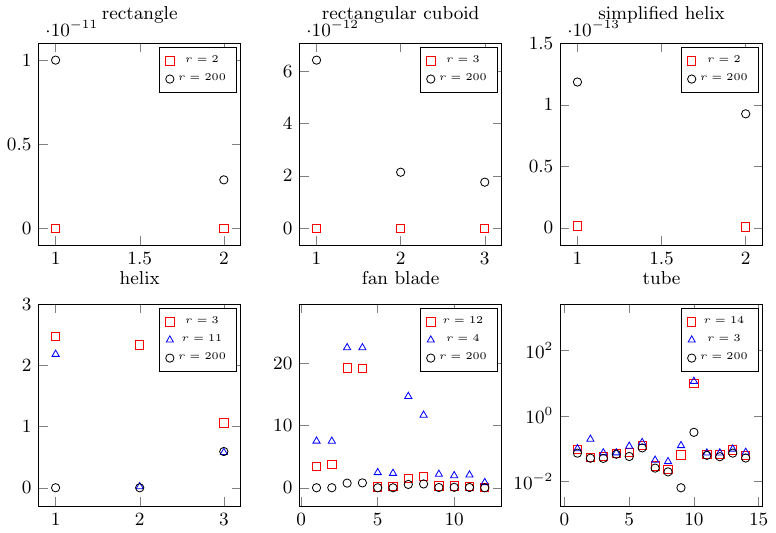}

	\caption{\label{fig:parameter_estimtion_mean_results}
        The mean of the absolute difference between the estimated and original parameters for each parameter for different numbers of used eigenvectors $r$.
        The red square denotes that the number of eigenvectors equals the number of design parameters, $r = k$. 
        With the blue triangle, we denote the number of eigenvectors equals the number of eigenvalues that stores $95\%$ of the total information, $r = t_{0.95}$. 
        And finally, the black circle denotes the use of 200 eigenvectors, $r=200$. 
        For classes, where $k = t_{0.95}$ only $r=k$ is presented. 
        The tube geometry class is logarithmically scaled, because the 10th parameter is much larger than the other ones. 
        In the first row, the classes from the first set are presented.
        The second row shows the geometry classes from the second set. 
        }
\end{figure}

At this point, the eigenvalues, eigenvectors $V$, and mean geometries $\mu_X$ for each geometry class are learned with the help of the PCA.
Now, with the help of the left equation of \eqref{eq:connection_eq_our_approach}, the mixing matrix $Z$ will be computed. 
And together with the mean vector of the parameter $\mu_p$, the matrix $H$, that stores the vectors of the parameter, is determined.
Now, Equation \eqref{eq:parameter_estimation_our_approach} can be employed to estimate the parameter for an arbitrarily transformed geometry $\vec{x}$ from the test dataset. 
The estimated parameter $\vec{p}_{\mathrm{e}}$ can then be compared to the original generating parameters $\vec{p}$. 
This error $\vec{e}$ is computed by the absolute difference between the estimated and original parameters, $\vec{\mathrm{e}} =  \lvert \vec{p}_{\mathrm{e}} - \vec{p} \rvert \in \mathbb{R}^k$.
Doing this for all 200 geometries within the test dataset, will result in 200 $\vec{\mathrm{e}}$ vectors.
The mean error $\vec{\mu}_{\mathrm{e}}$, is therefore the mean of all error vectors of a certain geometry class. 
These mean errors for the different geometry classes are shown in Figure \ref{fig:parameter_estimtion_mean_results}. 

During the computation, $Z$ we have the choice to vary the number $r$ of used eigenvectors. 
Like discussed earlier, it is not necessary to use all of them, since $t$ eigenvalues store enough information. 
This choice transfers to the parameter estimation, and the results for the different numbers of used eigenvectors $r$ can be seen in Figure \ref{fig:parameter_estimtion_mean_results}, too.
The first choice for $r$ is the number of generating parameters $k$ for the corresponding geometry class, $r = k$.
This choice is indicated via red squares, within the plots. 
The second choice is $t_{0.95}$ for a ratio of $\mathrm{CRV}_t = 0.95$, see Table \ref{tab:r_test_results}, $r = t_{0.95}$. 
The blue triangles indicate this choice of $r$. 
Since the value $t$ is equal to $k$ for some geometry classes, the blue triangles are omitted. 
The third choice is $r = 200$, as a high number, where you should include much more information as needed, and it is denoted with black circles. 
In the first row, of Figure \ref{fig:parameter_estimtion_mean_results} the geometry classes from the first set are presented. These are the geometry classes, where the class did not change during the centering process.
The second row, stores the plots of the classes in the second set. 

\subsubsection*{Discussion:}
There are two key observations from Figure \ref{fig:parameter_estimtion_mean_results}.
The first observation aims at the range of the mean error $\vec{\mu}_{\mathrm{e}}$. 
For the geometry classes from the first set, $\vec{\mu}_{\mathrm{e}}$ has a range of $10^{-11}$ up to $10^{-13}$. 
For the geometry classes of the second set, the ranges are $10^{1}$ up to $10^{2}$. 
The second observation is, that more eigenvectors do not lead to better estimations. 
Looking at the mean error for the different $r$ values and classes within the first set, it is noticeable that the mean error is higher for $r=200$ used eigenvectors. 
The black circles are over the red squares in the first row of Figure \ref{fig:parameter_estimtion_mean_results}.
For the geometry classes from the second set, second row of Figure \ref{fig:parameter_estimtion_mean_results}, a contradicting observation can be made.
The black circles are almost everywhere closest to zero. 

The visual investigation shows, that the parameter estimation for geometries that do not change the geometry class during the PCA process, is superior to the classes from the second set. 
Another point is, that if the geometry classes are changed, the parameter estimation can be improved, by using more eigenvectors.

\section{Conclusion\label{sec:Conclusion}}

In this work, we analyzed, by defining instructive examples, what is happening with geometry classes within the PCA.
We showed that for some geometry classes, a change of class occurs at the centering step in the PCA. 
Thus, the geometry classes were separated into two sets, the first set with the classes that do not change and the second set with classes that do change during the PCA. 
Further investigations with these two sets of geometry classes showed, that the eigenvectors of classes from the first set have more interpretable knowledge representation, than the eigenvectors from elements of the second set. 
The parameter estimation showed, that the eigenvectors of the classes from the first set are more suitable for this task, than the eigenvectors of the classes from the second set with the low interpretable knowledge representation. 

From our investigation, we conclude that the change of the geometry class may act like a qualitative criterion for the estimation of generating parameters from spatial properties.
Since, this property is difficult to make quantitative, another way of transforming a 3-D geometry into  a vector could be elaborated. 
Future work may focus on the task of creating an injective mapping, that maps the generating parameter and the spatial geometry into a single random vector.
This could be done with, e.g., a decoder-encoder network.

\subsubsection{Acknowledgments.} This work is part of the VIT VI project and is funded by the European Union and the state of Brandenburg, with the grand number 85062876. 
Furthermore, we would like to thank Carsten F\"utterer from Friendship Systems, and Maren Fanter from Rolls-Royce Germany for providing data and productive discussions.

\section*{Appendix}

\begin{figure}
	\centering
	\includegraphics[width = \linewidth]{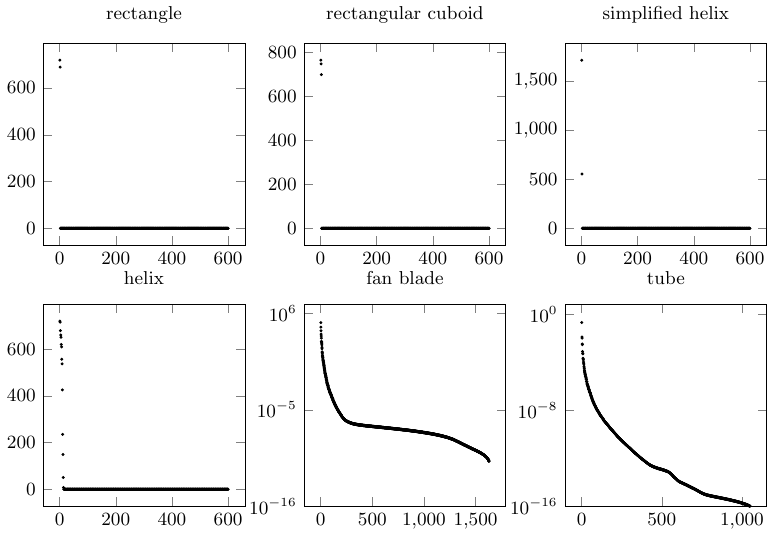}
	\caption{ \label{fig:eig_val_all_geometries}
        The eigenvalues for all geometry classes. In the top row, the geometry classes from the first set are presented. The second row, shows the classes from the second set. 
        For the fan blade and tube class, the $y$ axis is logarithmically scaled and limited downward at $10^{-16}$.}
\end{figure}

\bibliographystyle{bibtex/splncs_srt}
\bibliography{references}

@article{Bouwmans2018,
  author    = {Bouwmans, Thierry and Javed, Sajid and Zhang, Hongyang and Lin, Zhouchen and Otazo, Ricardo},
  journal   = {Proceedings of the IEEE},
  title     = {{On the Applications of Robust PCA in Image and Video Processing}},
  year      = {2018},
  issn      = {1558-2256},
  month     = aug,
  number    = {8},
  pages     = {1427--1457},
  volume    = {106},
  doi       = {10.1109/jproc.2018.2853589},
  publisher = {Institute of Electrical and Electronics Engineers (IEEE)}
}

@inproceedings{Chen2017,
  author    = {Chen, Jiachen and Jenkins, W. Kenneth},
  booktitle = {2017 IEEE 60\textsuperscript{th} International Midwest Symposium on Circuits and Systems (MWSCAS)},
  title     = {{Facial recognition with PCA and machine learning methods}},
  year      = {2017},
  month     = aug,
  publisher = {IEEE},
  doi       = {10.1109/mwscas.2017.8053088}
}

@article{Hotelling1933,
  author    = {Harold Hotelling},
  journal   = {Journal of Educational Psychology},
  title     = {Analysis of a complex of statistical variables into principal components.},
  year      = {1933},
  month     = sep,
  number    = {6},
  pages     = {417--441},
  volume    = {24},
  comment   = {This paper by Harold Hotelling is often cited as the first paper to use the term "principal component analysis (PCA)" and to provide a clear exposition of the method.},
  doi       = {10.1037/h0071325},
  publisher = {American Psychological Association ({APA})}
}

@inbook{Howley2005,
  author    = {Howley, Tom and Madden, Michael G. and O{\textquoteright}Connell, Marie-Louise and Ryder, Alan G.},
  pages     = {209--222},
  publisher = {Springer London},
  title     = {{The Effect of Principal Component Analysis on Machine Learning Accuracy with High Dimensional Spectral Data}},
  year      = {2005},
  isbn      = {9781846282232},
  booktitle = {Applications and Innovations in Intelligent Systems XIII},
  doi       = {10.1007/1-84628-224-1_16}
}

@book{Jolliffe2002,
  author    = {I. T. Jolliffe},
  publisher = {Springer-Verlag},
  title     = {{Principal Component Analysis}},
  year      = {2002},
  edition   = {2},
  isbn      = {978-0-387-95442-4},
  series    = {Springer Series in Statistics},
  comment   = {This book provides a comprehensive introduction to PCA, covering both the theory and applications of the method.},
  date      = {2002-10-01},
  doi       = {10.1007/b98835},
  ean       = {9780387954424},
  pagetotal = {488}
}

@article{Le2020,
  author    = {Le, Thai-Ha and Le, Ha-Chi and Taghizadeh-Hesary, Farhad},
  journal   = {Finance Research Letters},
  title     = {{Does financial inclusion impact CO2 emissions? Evidence from Asia}},
  year      = {2020},
  issn      = {1544-6123},
  pages     = {101451},
  volume    = {34},
  doi       = {10.1016/j.frl.2020.101451},
  publisher = {Elsevier BV}
}

@article{McVean2009,
  author    = {McVean, Gil},
  journal   = {PLoS Genetics},
  title     = {{A Genealogical Interpretation of Principal Components Analysis}},
  year      = {2009},
  issn      = {1553-7404},
  number    = {10},
  pages     = {e1000686},
  volume    = {5},
  doi       = {10.1371/journal.pgen.1000686},
  editor    = {Przeworski, Molly},
  publisher = {Public Library of Science (PLoS)}
}

@article{Nobre2019,
  author    = {Nobre, João and Neves, Rui Ferreira},
  journal   = {Expert Systems with Applications},
  title     = {{Combining Principal Component Analysis, Discrete Wavelet Transform and XGBoost to trade in the financial markets}},
  year      = {2019},
  issn      = {0957-4174},
  pages     = {181--194},
  volume    = {125},
  doi       = {10.1016/j.eswa.2019.01.083},
  publisher = {Elsevier BV}
}

@inproceedings{Patil2011,
  author    = {Patil, Ujwala and Mudengudi, Uma},
  booktitle = {2011 International Conference on Image Information Processing},
  title     = {{Image fusion using hierarchical PCA.}},
  year      = {2011},
  month     = nov,
  publisher = {IEEE},
  doi       = {10.1109/iciip.2011.6108966}
}

@article{Pearson1901,
  author    = {Karl Pearson},
  journal   = {Philosophical Magazine Series 1},
  title     = {{LIII. On lines and planes of closest fit to systems of points in space}},
  year      = {1901},
  pages     = {559--572},
  volume    = {2},
  comment   = {this is the original paper in which PCA was introduced by Karl Pearson in 1901.}
}

@article{Reich2008,
  author    = {Reich, David and Price, Alkes L. and Patterson, Nick},
  journal   = {Nature Genetics},
  title     = {{Principal component analysis of genetic data}},
  year      = {2008},
  issn      = {1546-1718},
  month     = may,
  number    = {5},
  pages     = {491--492},
  volume    = {40},
  doi       = {10.1038/ng0508-491},
  publisher = {Springer Science and Business Media LLC}
}

@Article{Serani2023,
  author           = {Serani, Andrea and Diez, Matteo},
  journal          = {Computer Methods in Applied Mechanics and Engineering},
  title            = {Parametric Model Embedding},
  year             = {2023},
  issn             = {0045-7825},
  month            = feb,
  pages            = {115776},
  volume           = {404},
  archiveprefix    = {arXiv},
  copyright        = {arXiv.org perpetual, non-exclusive license},
  creationdate     = {2025-10-27T11:27:02},
  date             = {2022-04-11},
  doi              = {10.1016/j.cma.2022.115776},
  eprint           = {2204.05371},
  modificationdate = {2025-10-27T11:27:20},
  primaryclass     = {math.OC},
  publisher        = {Elsevier BV},
}

@Article{Serani2025,
  author           = {Serani, Andrea and Palma, Giorgio and Wackers, Jeroen and Quagliarella, Domenico and Gaggero, Stefano and Diez, Matteo},
  journal          = {Engineering with Computers},
  title            = {{Extending parametric model embedding with physical information for design-space dimensionality reduction in shape optimization}},
  year             = {2025},
  issn             = {1435-5663},
  month            = oct,
  number           = {6},
  pages            = {4643--4663},
  volume           = {41},
  creationdate     = {2026-01-13T10:32:14},
  doi              = {10.1007/s00366-025-02211-2},
  modificationdate = {2026-01-13T10:32:33},
  publisher        = {Springer Science and Business Media LLC},
}

@Article{Phuong2025,
  author           = {Phuong, Nguyen Dong and Subbiah, Nanthakumar Srivilliputtur and Jin, Yabin and Zhuang, Xiaoying},
  journal          = {Computer Modeling in Engineering \& Sciences},
  title            = {Deep Learning-Based Inverse Design: Exploring Latent Space Information for Geometric Structure Optimization},
  year             = {2025},
  issn             = {1526-1506},
  number           = {1},
  pages            = {263--303},
  volume           = {145},
  creationdate     = {2026-01-13T10:37:52},
  doi              = {10.32604/cmes.2025.067100},
  modificationdate = {2026-01-13T10:38:03},
  publisher        = {Tech Science Press},
}

@Article{Penrose1955,
  author           = {Penrose, R.},
  journal          = {Mathematical Proceedings of the Cambridge Philosophical Society},
  title            = {A generalized inverse for matrices},
  year             = {1955},
  issn             = {1469-8064},
  month            = jul,
  number           = {3},
  pages            = {406--413},
  volume           = {51},
  creationdate     = {2026-01-22T10:44:08},
  doi              = {10.1017/s0305004100030401},
  modificationdate = {2026-01-22T10:44:10},
  publisher        = {Cambridge University Press (CUP)},
}

@InProceedings{Bjerhammar1951,
  author           = {Arne Bjerhammar},
  title            = {Application of calculus of matrices to method of least squares: with special reference to geodetic calculations},
  year             = {1951},
  modificationdate = {2026-01-22T10:47:25},
  url              = {https://api.semanticscholar.org/CorpusID:118134976},
}

@Article{Moore1920,
  author           = {Eliakim Hastings Moore},
  journal          = {Bulletin of the American Mathematical Society},
  title            = {On the reciprocal of the general algebraic matrix},
  year             = {1920},
  creationdate     = {2026-01-22T11:07:13},
  modificationdate = {2026-01-22T11:08:27},
}

@Article{Diez2015,
  author           = {Diez, Matteo and Campana, Emilio F. and Stern, Frederick},
  journal          = {Computer Methods in Applied Mechanics and Engineering},
  title            = {Design-space dimensionality reduction in shape optimization by Karhunen--Lo{\`{e}}ve expansion},
  year             = {2015},
  issn             = {0045-7825},
  month            = jan,
  pages            = {1525--1544},
  volume           = {283},
  creationdate     = {2026-01-29T15:40:48},
  doi              = {10.1016/j.cma.2014.10.042},
  modificationdate = {2026-01-29T15:40:58},
  publisher        = {Elsevier BV},
}

@Article{Cinquegrana2018,
  author           = {Cinquegrana, Davide and Iuliano, Emiliano},
  journal          = {Computers \& Fluids},
  title            = {Investigation of adaptive design variables bounds in dimensionality reduction for aerodynamic shape optimization},
  year             = {2018},
  issn             = {0045-7930},
  month            = sep,
  pages            = {89--109},
  volume           = {174},
  creationdate     = {2026-01-29T15:45:53},
  doi              = {10.1016/j.compfluid.2018.07.012},
  modificationdate = {2026-01-29T15:45:53},
  publisher        = {Elsevier BV},
}

@Article{Yasong2018,
  author           = {QIU, Yasong and BAI, Junqiang and LIU, Nan and WANG, Chen},
  journal          = {Chinese Journal of Aeronautics},
  title            = {Global aerodynamic design optimization based on data dimensionality reduction},
  year             = {2018},
  issn             = {1000-9361},
  month            = apr,
  number           = {4},
  pages            = {643--659},
  volume           = {31},
  creationdate     = {2026-01-29T15:47:16},
  doi              = {10.1016/j.cja.2018.02.005},
  modificationdate = {2026-01-29T15:47:18},
  publisher        = {Elsevier BV},
}

@Article{Harries2021,
  author           = {Harries, Stefan and Uharek, Sebastian},
  journal          = {Journal of Marine Science and Engineering},
  title            = {Application of Radial Basis Functions for Partially-Parametric Modeling and Principal Component Analysis for Faster Hydrodynamic Optimization of a Catamaran},
  year             = {2021},
  issn             = {2077-1312},
  month            = sep,
  number           = {10},
  pages            = {1069},
  volume           = {9},
  creationdate     = {2026-01-29T15:52:35},
  doi              = {10.3390/jmse9101069},
  modificationdate = {2026-01-29T15:52:35},
  publisher        = {MDPI AG},
}

\end{document}